\documentclass[10pt, a4paper]{article}
\pdfoutput=1
\usepackage[final]{lrec-coling2024}
\makeatletter
\def\@mb@citenamelist{cite,citep,citet,citealp,citealt,citepalias,citetalias}
\makeatother
\newcites{languageresource}{~}

\usepackage{graphicx}
\usepackage{tabularx}
\usepackage{booktabs}
\usepackage{xcolor}
\usepackage{hyperref}
 \definecolor{darkblue}{rgb}{0, 0, 0.5}
  \hypersetup{colorlinks=true, citecolor=darkblue, linkcolor=darkblue, urlcolor=darkblue}

\title{SPICED: News Similarity Detection Dataset with Multiple Topics and Complexity Levels} 

\name{Elena Shushkevich$^{\ast}$, Manuel V. Loureiro$^{\dagger}$,\\
    {\bf \large Long Mai$^{\ddagger}$, Steven Derby$^{\dagger}$, Tri Kurniawan Wijaya$^{\dagger}$}} 

\address{
    $^{\ast}$Technological University Dublin \\ elena.n.shushkevich@gmail.com \\ \\
    $^{\dagger}$Huawei Ireland Research Centre \\ manuel.loureiro@huawei.com, steven.derby@huawei-partners.com, \\ tri.kurniawan.wijaya@huawei.com \\ \\
    $^{\ddagger}$University College Dublin \\ long.mai@ucdconnect.ie
}

\abstract{
The proliferation of news media outlets has increased the demand for intelligent systems capable of detecting redundant information in news articles in order to enhance user experience. However, the heterogeneous nature of news can lead to spurious findings in these systems: Simple heuristics such as whether a pair of news are both about politics can provide strong but deceptive downstream performance. 
Segmenting news similarity datasets into topics improves the training of these models by forcing them to learn how to distinguish salient characteristics under more narrow domains. However, this requires the existence of topic-specific datasets, which are currently lacking.
In this article, we propose a novel dataset of similar news, SPICED, which includes seven topics: Crime \& Law, Culture \& Entertainment, Disasters \& Accidents, Economy \& Business, Politics \& Conflicts, Science \& Technology, and Sports.  
Futhermore, we present four different levels of complexity, specifically designed for news similarity detection task.
We benchmarked the created datasets using MinHash, BERT, SBERT, and SimCSE models.
    \\ \newline \Keywords{dataset, news similarity, text similarity detection}
}

\begin{document}

\maketitleabstract

\section{Introduction}

\begin{table*}[ht]
    \centering
    {
        \begin{tabularx}{\textwidth}{>{\raggedright\arraybackslash}Xrrrrrrr}
            \toprule
            \textbf{Topics}    & \textbf{CL} & \textbf{CE} & \textbf{DA} & \textbf{EB} & \textbf{PC} & \textbf{ST} & \textbf{SP} \\
            \midrule \specialrule{.1em}{.05em}{.05em}
            \multicolumn{4}{l}{\textbf{Document statistics}}                                                                     \\
            \midrule
            \# Webpages        & 4,419       & 2,129       & 2,757       & 2,320       & 7,675       & 2,064       & 2,423       \\
            \# Source articles & 7,495       & 3,759       & 4,716       & 3,881       & 14,075      & 3,681       & 3,738       \\
            \midrule \specialrule{.1em}{.05em}{.05em}
            \multicolumn{4}{l}{\textbf{Words per source}}                                                                        \\
            \midrule
            Mean               & 606.9       & 518.4       & 544.7       & 605.1       & 629.6       & 662.1       & 579.3       \\
            Median             & 553         & 409         & 487         & 519         & 563         & 583         & 472         \\
            Minimum            & 34          & 26          & 37          & 31          & 34          & 42          & 43          \\
            Maximum            & 2,420       & 2,974       & 2,918       & 3,663       & 2,514       & 3,092       & 2,200       \\
            \bottomrule
        \end{tabularx}}
    \caption{\label{table:statistics} The number of collected documents per topics and the mean, median, minimum, and maximum word counts per article for the seven topics: Crime \& Law (CL), Culture \& Entertainment (CE), Disasters \& Accidents (DA), Economy \& Business (EB), Politics \& Conflicts (PC), Science \& Technology (ST), Sports (SP) topics.}
\end{table*}
The rise of the internet has ushered in an era of unprecedented growth in online publishing, resulting in a deluge of news content. In this digital landscape, users often find themselves investing significant time and effort in navigating a multitude of articles, all covering the same events, especially within news aggregator platforms. This abundance of information can present a formidable challenge, as it becomes increasingly difficult to discern and access the specific and relevant content that users seek amidst the vast sea of news articles.

Publicly available training resources are scarce for developing systems for similar news article detection. Existing semantic textual similarity (STS) datasets are not suitable for news similarity detection, as they are specific to a single topic, such as MedSTS \cite{wang2020medsts} and CORD19STS \cite{guo2020cord19sts}.
However, news similarity detection is inherently influenced by the high degree of heterogeneity in news content and structure, which follows a well-understood taxonomy based on news categories or genres. These categories contain salient overlapping semantic features which directly affect how difficult it is to compare news articles, which can result in systems which quickly learn unproductive ruled-based heuristics rather than the complex linguistic structure relating these text documents \cite{wu2021esimcse}. For example, sports news generally contains more distinctive features and less ambiguity than political news. Therefore, we need to compare news across different categories as well as within the same category, to better assess the performance of different models on different levels of similarity.
To this end, high-quality datasets are crucial for improving news similarity detection in complex cases, where similarity within the same topic is harder to discern than between unrelated topics.

In this work, we propose SPICED (\textbf{S}cience, \textbf{S}ports, \textbf{P}olitics, \textbf{C}rime, \textbf{C}ulture, \textbf{E}conomy, \textbf{D}isasters), a multi-topic dataset addressing the mentioned problems. It includes Crime \& Law, Culture \& Entertainment, Disasters \& Accidents, Economy \& Business, Politics \& Conflicts, Science \& Technology, and Sports topics. The full dataset is publicly available\footnote{You can download the dataset at \url{https://zenodo.org/record/8044777}}.

By utilizing the original dataset, we propose four distinct approaches for creating pairs in the context of the news similarity detection task. Each approach offers a unique combination of true similar and false similar news pairs.

Our contributions are as follows:
\begin{itemize}
    \item We provide an original dataset of 977 similar news pairs in English (1,954 news articles), devoted to the seven different popular news topics.
    \item We provide 32 datasets, all derived from an original gold-standard dataset. These datasets represent four different complexity levels for creating news pairs within the context of both single-topic and multi-topic similar news detection.
    \item We benchmark these created datasets using four algorithms for STS tasks which are prevalent in the literature: MinHash, BERT, SBERT, and SimCSE.
\end{itemize}

\section{Related Work}
Finding a suitable dataset for the news similarity detection task is challenging, as we need a dataset that uses categorized news article pairs to measure news similarity, but to the best of our knowledge there is no such dataset. However, there are still other datasets that are valuable for this task.

SemEval-2022 Task 8 is a multilingual news article similarity dataset of approximately 10,000 news article pairs encompassing 18 combinations of 10 languages \cite{chen2022semeval}. The similarity scores result from the averages of multiple human annotators using a four-point Likert scale over seven dimensions measuring geographic, temporal, and narrative similarities. Notice that this dataset does not contain any information regarding the classification of news articles over some taxonomy.

SentEval is a toolkit to evaluate the quality of universal sentence representations over various tasks, such as binary and multi-class classification, natural language inference, and sentence similarity \cite{conneau2018senteval} allowing the evaluation of sentence embeddings as features for many semantic textual similarity downstream tasks \cite{agirre2012semeval,agirre2013sem, agirre2014semeval,agirre2015semeval,agirre2016semeval,cer2017semeval}.  The data for these datasets were collected from news articles, forum discussions, news conversations, headlines, image and video descriptions, and were labeled with a similarity score between 0 and 5. The target of the tasks was to evaluate the distance between two sentences using cosine distance \cite{rahutomo2012semantic}. SentEval tasks included some subtasks and reported both the average and the weighted average (by number of samples in each subtask) of the Pearson \cite{freedman2020statistics} and Spearman \cite{zar2005spearman} correlations.

Entailment and semantic relatedness detection tasks can also be useful in the context of similar news detection. In this area, it is important to mention the SICK dataset \cite{marelli2014sick}, which consists of about 10,000 English sentence pairs, each of which is annotated for both entailment task – SICK-E (with three possible labels: entailment, contradiction, and neutral) and relatedness detection task – SICK-R (with 5-point rating scale as gold score).

In addition, the paraphrase detection task is comparable with the similar news detection task, where we highlight the Microsoft Research Paraphrase Corpus (MRPC) \cite{dolan2004unsupervised,dolan-brockett-2005-automatically}, for which the goal is to identify if two sentences are variations based on synonymy and local syntactic changes. MRPC is a monolingual dataset presenting 5,801 naturally occurring paraphrase pairs extracted from over 13 million sentence pairs collected from the World Wide Web using heuristic techniques and an automatic classifier, with 67\% of the paraphrase pairs deemed semantically equivalent by human annotators.

\begin{table*}[ht]
    \centering
    \begin{tabularx}{\textwidth}{>{\raggedright\arraybackslash}Xrrrrrrr}
        \toprule
        \textbf{Topics}     & \textbf{CL} & \textbf{CE} & \textbf{DA} & \textbf{EB} & \textbf{PC} & \textbf{ST} & \textbf{SP} \\
        \midrule \specialrule{.1em}{.05em}{.05em}
        \multicolumn{4}{l}{\textbf{Filters}}                                                                                  \\
        \midrule
        SimHash             & 76,996      & 8,672       & 24,015      & 30,291      & 123,791     & 8,916       & 14,954      \\
        \shortstack{Source of the same                                                                                        \\Wikinews page} & 511 & 259 & 316 & 312 & 822 & 273 & 334\\
        SBERT               & 501         & 230         & 300         & 279         & 779         & 249         & 318         \\
        Experts’ annotation & 238         & 95          & 137         & 120         & 361         & 136         & 94          \\
        Duplicates removal  & 192         & 90          & 124         & 107         & 259         & 111         & 94          \\
        \bottomrule
    \end{tabularx}
    \caption{\label{table:filters}
        The number of similar pairs after each sequential filtering step for Crime and Law (CL), Culture and Entertainment (CE), Disasters and Accidents (DA), Economy and Business (EB), Politics and Conflicts (PC), Science and Technology (ST), Sports (SP) topics. As we move down through each filtering step the number of articles is reduced to build our gold-standard dataset.}
\end{table*}

\section{Dataset creation}

This section is dedicated to discussing the process of creating the news article similarity dataset. This dataset is comprised of paired articles, each associated with a binary similarity label, serving as a fundamental resource for our research.

\subsection{Collecting News Articles}

\emph{WikiNews}\footnote{\url{https://en.wikinews.org/wiki/Main_Page}}, a collaborative journalism initiative under the Wikimedia Foundation, adheres to specific guidelines\footnote{\url{https://en.wikinews.org/wiki/Wikinews:Pillars_of_writing}} for news article creation. These guidelines stipulate that news articles must be categorized by topic and substantiated by a minimum of two independent and authoritative sources. Our selection process exclusively considered sources with valid and accessible URLs. Given that these sources collectively cover the same news events and provide substantial information, they can be reasonably deemed similar. This alignment in content served as the basis for the development of the proposed news article similarity dataset.

In the month of April 2022, our data collection efforts were directed towards gathering WikiNews articles, employing the utility of \emph{BeautifulSoup}\footnote{\url{https://www.crummy.com/software/BeautifulSoup/bs4/doc/}}, a versatile web scraping tool. Specifically, we focused on harvesting articles belonging to the seven most populous categories, namely: Crime \& Law, Culture, Disasters \& Accidents, Economy \& Business, Politics \& Conflicts, Science \& Technology, and Sports. The statistics detailing the outcome of this data collection process are documented in Table \ref{table:statistics}.

\subsection{Measuring Similar News}
We begin by utilizing baseline similarity models in order to query news article similarity, as a way to supplement our raw data. Because there are a combinatorically impractical number of possible document pairs, we offload some of this work to these oracles, as they enable us to efficiently identify suitable examples.

The \emph{SimHash} algorithm\footnote{\url{https://github.com/1e0ng/simhash}} \cite{sadowski2007simhash} is employed to identify pairs of news articles with high similarity. Determining similarity or dissimilarity is based on a threshold specified within the SimHash implementation. Subsequently, a validation process is conducted to ensure that both news articles in a pair originate from the same WikiNews webpage.

Next, for the subset of similar news articles (according to the SimHash filtering step) originating from the same WikiNews webpage, we utilize the transformer-based model \emph{SBERT}\cite{reimers2019sentence}, specifically the {paraphrase-multilingual-mpnet-base-v2}\footnote{We use \href{https://huggingface.co/sentence-transformers/paraphrase-multilingual-mpnet-base-v2}{paraphrase-multilingual-mpnet-base-v2}.} model, to identify the most similar news articles within the dataset. The approach of creating SimHash pairs separately for each topic is applied consistently.

\subsection{Dataset Annotation}
Experts review and assess rudimentary approximated news pairs to gather suitable samples for our final gold-standard annotations. Two annotators evaluate all proposed pairs and reach a consensus on their similarity. The annotators are two PhD students. Before commencing work on the dataset, these annotators, in collaboration with the authors, participated in joint reviews of several external news articles to establish a shared understanding of the task at hand. They resolve discrepancies through discussion to determine whether to retain or discard the pair. Disagreements on the annotations were addressed on a case-by-case basis, with each annotator presenting their perspective to reach a consensus.

For experts annotation, we define the following criteria that any similar pair of news articles must satisfy:
\begin{enumerate}
    \item Both news articles in a pair must be about the same topic and event (for example, topic – sports, event – UEFA Champions League final);
    \item Both news articles should have similar lengths to avoid information asymmetry, where one article contains significantly more information than the other;
    \item Opinion articles, prone to biases, should be excluded from similar news classifications. Similar news should be factual and not influenced by the authors' interpretations;
    \item Any numerical values cited in the articles should be consistent. For example, if one article mentions 10 road accident victims and its pair states "more than 8 people," they should still be considered similar;
    \item The time of publication must be close. News articles discussing the same event but published at significantly different times are considered dissimilar.
\end{enumerate}
The last step of the filtering is to delete duplicate pairs, which can appear in cases when news articles are devoted to several topics at once. For example, a news article can be both about politics and economics. As we strive for the most balanced dataset, we remove the pair from the topic with the bigger number of samples.

In the Table\ref{table:examples}, some examples of similar news are presented. All four news examples are about events in Australia, but the first pair pertains to Sports topic, while the second pair relates to Disasters \& Accidents topic.

\begin{table*}[ht]
    \centering
    \begin{tabularx}{\textwidth}{>{\raggedright\arraybackslash}ll}
        \toprule
        \textbf{News 1}                                         & \textbf{News 2}                                \\
        \hline
                                                                &                                                \\
        Sydney FC penalised for contract breach                 & Sydney FC lose points                          \\
        Sydney FC is in greater danger of missing the           & Sydney FC will have three competition points   \\
        A-League play-offs after being hit with                 & deducted and be fined \$129,000 after          \\
        a deduction of three competition points for             & Football Federation Australia (FFA) today      \\
        breaching Player Contracting Regulations.               & found the club guilty of breaching the         \\
        Football Federation Australia has also decided          & A-League's Player Contracting Regulations.     \\
        to fine the club \$129,000. Sydney won't lose           & Sydney won't lose the points                   \\
        the points immediately and has seven days to appeal.    & immediately and has seven days to appeal.      \\
                                                                & If the club chooses not to appeal, the points  \\
                                                                & will be deducted next Friday.                  \\
                                                                &                                                \\
        \hline
                                                                &                                                \\
        Great Barrier Reef oil disaster fear from stricken ship & Australia warns stranded Chinese ship          \\
        The Shen Neng 1 was nine miles (15 km)                  & could break up SYDNEY (Reuters)                \\
        outside the shipping lane A Chinese ship is             & A stranded Chinese bulk coal carrier leaking   \\
        in danger of breaking up after running aground          & oil into the sea around Australia’s Great      \\
        off north-east Australia, sparking fears                & Barrier Reef is in danger of breaking up       \\
        of a major oil spill into the Great Barrier Reef.       & and damaging the reef, government officials    \\
        The Shen Neng 1, carrying 950 tonnes                    & said on Sunday. Oil is seen next to the        \\
        of oil, ran aground 70km (43 miles) off the east        & 230-metre (754-ft) Chinese bulk coal carrier   \\
        coast of Great Keppel Island. Some                      & Shen Neng I, about 70 km (43 miles) east       \\
        oil has already leaked and there are fears              & of Great Keppel Island April 4, 2010. The      \\
        the coal-carrier may split into parts, causing          & 230-meter (754-ft) Shen Neng I was on          \\
        a greater spillage. The Australian authorities          & its way to China when it ran aground on a      \\
        say the ship was in a protected area, well              & shoal on Saturday. It had 950 tonnes of oil    \\
        outside the normal shipping channels.                   & on board and officials said patches of oil     \\
        Chemical dispersants are being used to prevent          & had been spotted in the water early on Sunday,
        \\
        the spill threatening the World                         & but no major leak...                           \\
        Heritage-listed marine reserve...                       &                                                \\
                                                                &                                                \\
        \bottomrule
    \end{tabularx}
    \caption{\label{table:examples} Examples of similar news pairs.}
\end{table*}

\subsection{Statistics}
In Table~\ref{table:filters} we present the number of similar news pairs at the end of each filtering step, including SimHash, the confirmation that both news in the pair is the sources of the same page, SBERT, experts’ annotation, and duplicates removal. To sum up, after the news scraping we implemented sequentially four steps to confirm the news pairs' similarity, including both ML approaches (SimHash, SBERT) and manual checks (2 experts annotations).
The number of similar pairs after the final step - duplicates removal - is the final number of similar pairs for each topic.
We divided the datasets with a ratio of 70:30 for training and test datasets, and we used this ratio for the pairs from the general dataset for each complexity level creation.
As shown in Table~\ref{table:statistics}, the average number of tokens in the article ranges from 518.4 to 662.1 tokens, while the maximum can reach 3663 tokens on the EB topic.
The final dataset contains 977 similar pairs. The dataset contains a much larger number of sentences than other similar datasets, which makes the task of computing their similarity more challenging. The dataset also covers a wide range of token lengths. This is beneficial for developing similarity models that can handle news articles of varying lengths.

\begin{table*}[ht]
    \normalsize
    \centering
    \begin{tabularx}{\textwidth}{>{\raggedright\arraybackslash}Xrr}
        \toprule
        \textbf{Model}                & Train   & Test    \\
        \midrule \specialrule{.1em}{.05em}{.05em}
        {\textbf{Inter Topic}}                            \\
        \midrule
        All                           & 767,587 & 148,382 \\
        \midrule \specialrule{.1em}{.05em}{.05em}
        \multicolumn{3}{l}{\textbf{Intra Topic}}          \\
        \midrule
        Crime \& Law (CL)             & 33,678  & 5,770   \\
        Culture \& Entertainment (CE) & 5,526   & 640     \\
        Disaster \& Accidents (DA)    & 12,606  & 1,950   \\
        Economy \& Business (EB)      & 8,778   & 1,245   \\
        Politics \& Conflict (PC)     & 63,241  & 11,190  \\
        Science \& Technology (ST)    & 9,681   & 1,378   \\
        Sporting Activities (SP)      & 6,285   & 753     \\
        \midrule \specialrule{.1em}{.05em}{.05em}
        \multicolumn{3}{l}{\textbf{Hard Examples}}        \\
        \midrule
        Crime \& Law (CL)             & 2,234   & 958     \\
        Culture \& Entertainment (CE) & 2,162   & 928     \\
        Disaster \& Accidents (DA)    & 2,186   & 938     \\
        Economy \& Business (EB)      & 2,174   & 933     \\
        Politics \& Conflict (PC)     & 2,281   & 978     \\
        Science \& Technology (ST)    & 2,177   & 934     \\
        Sporting Activities (SP)      & 2,165   & 929     \\
        \midrule \specialrule{.1em}{.05em}{.05em}
        \multicolumn{3}{l}{\textbf{Combined}}             \\
        \midrule
        All                           & 921,403 & 177,310 \\
        \bottomrule
    \end{tabularx}
    \caption{\label{tab3}
        The number of train and test instances (news pairs) for each topic and complexity level.}
\end{table*}

\begin{table*}[h]
    \normalsize
    \centering
    \begin{tabularx}{\textwidth}{>{\raggedright\arraybackslash}Xrrrr}
        \toprule
        \textbf{Model}                       & MinHash        & BERT           & SBERT          & SimCSE         \\
        \midrule \specialrule{.1em}{.05em}{.05em}
        \multicolumn{4}{l}{\textbf{Inter-Topic}}                                                                 \\
        \midrule
        \textit{F1-score}                    & \textit{0.707} & \textit{0.786} & \textit{0.920} & \textit{0.896} \\
        \midrule \specialrule{.1em}{.05em}{.05em}
        \multicolumn{4}{l}{\textbf{Intra-Topic}}                                                                 \\
        \midrule
        Crime \& Law (CL)                    & 0.816          & 0.851          & 0.957          & 0.957          \\
        Culture \& Entertainment (CE)        & 0.902          & 0.923          & 0.923          & 0.943          \\
        Disaster \& Accidents (DA)           & 0.742          & 0.853          & 0.935          & 0.853          \\
        Economy \& Business (EB)             & 0.678          & 0.828          & 0.899          & 0.937          \\
        Politics \& Conflict (PC)            & 0.650          & 0.776          & 0.911          & 0.875          \\
        Science \& Technology (ST)           & 0.690          & 0.824          & 0.921          & 0.847          \\
        Sporting Activities (SP)             & 0.840          & 0.840          & 0.982          & 0.816          \\
        \midrule
        \textit{Average F1-score}            & \textit{0.760} & \textit{0.842} & \textit{0.933} & \textit{0.890} \\
        \midrule \specialrule{.1em}{.05em}{.05em}
        \multicolumn{4}{l}{\textbf{Hard Examples}}                                                               \\
        \midrule
        Crime \& Law (CL)                    & 0.727          & 0.891          & 0.935          & 0.919          \\
        Cultu921403-re \& Entertainment (CE) & 0.833          & 0.906          & 0.902          & 0.943          \\
        Disaster \& Accidents (DA)           & 0.742          & 0.795          & 0.938          & 0.868          \\
        Economy \& Business (EB)             & 0.690          & 0.774          & 0.952          & 0.909          \\
        Politics \& Conflict (PC)            & 0.702          & 0.829          & 0.940          & 0.892          \\
        Science \& Technology (ST)           & 0.741          & 0.639          & 0.853          & 0.667          \\
        Sporting Activities (SP)             & 0.840          & 0.840          & 0.945          & 0.964          \\
        \midrule
        \textit{Average F1-score}            & \textit{0.754} & \textit{0.811} & \textit{0.924} & \textit{0.880} \\
        \midrule \specialrule{.1em}{.05em}{.05em}
        \multicolumn{4}{l}{\textbf{Combined}}                                                                    \\
        \midrule
        \textit{F1-score}                    & \textit{0.757} & \textit{0.799} & \textit{0.922} & \textit{0.875} \\
        \bottomrule
    \end{tabularx}
    \caption{\label{tab2}
        F1-scores of the experiments on various topics and complexity levels.
    }
\end{table*}

\subsection{Complexity Levels}
We present an additional contribution and novelty by
introducing several complexity levels
within our datasets.

For a more detailed analysis of the created dataset,
we divide it into four complexity levels,
which allows us to explore not only different news topics,
but also their degree of similarity.
Table~\ref{tab3} provides the number of training and test instances
for each dataset corresponding to each complexity levels.

\textbf{Inter-Topic.}
This set includes similar news pairs as positive samples and dissimilar news pairs from different topics as negative samples. This approach distinguishes dissimilar pairs when they belong to different topics.

\textbf{Intra-Topic.}
This set contains positive and negative pairs within the same topic, split into seven separate subsets corresponding to different topics. We also remove the challenging examples from the negative pairs, as they will belong to the next approach's datasets.

\textbf{Hard Examples.}
This set consists of all positive pairs and the 3,000 most similar negative pairs, according to SimHash, within each intra-topic.
The way the negative pairs are carefully chosen here makes it less obvious to distinguish similar news from the dissimilar ones.
To ensure that there is no overlap between the negative pairs in the intra-topic and hard examples sets,
we exclude these pairs from their corresponding intra-topic set. The results demonstrate that training a model on partitioned categories provides better improvements than hard-mining examples, though we note the number of hard examples are smaller.

\textbf{Combined.}
While the set of news pairs in the previous three complexity levels are disjoint,
this \emph{combined} set includes all (union) of the positive and negative news pairs from the previous sets.

Thus, we have our original gold-standard dataset containing 977 pairs that belong to a variety of categories (Crime \& Law, Culture \& Entertainment, Disasters \& Accidents, Economy \& Business, Politics \& Conflicts, Science \& Technology, and Sports), which consists of 1954 articles. From here, we split the dataset into 679 training and 298 test pairs and create labels by generating incorrect pairs. These pairs represent a total of 1358 training articles (and 596 testing articles), resulting in 921,403 pairs for training (and 177,310 pairs for testing), encompassing all combined examples. Inter-topic pairs are those that do not belong to the same category, while intra-topic pairs are those that do. Hard examples are pairs that belong to the same category but are considered difficult because they are incorrect yet highly similar to the base article.

\section{Benchmarking}
This section describes the models, experiments and benchmark results of our novel similarity news dataset.

\subsection{Pretrained Models}
\textbf{Minhash} is an efficient method for estimating set similarity using the Jaccard coefficient~\cite{broder1997resemblance}.
We used the \textit{snapy} library\footnote{\url{https://libraries.io/pypi/snapy}} to obtain a simple baseline for more complex algorithms.

\textbf{BERT} (Bidirectional Encoder Representations from Transformers) represents a classical approach employed in this study for obtaining embeddings of news articles~\cite{devlin2018bert}.
Subsequently, cosine similarities are calculated between these embeddings to gauge their semantic similarity. For this purpose, we leveraged the widely recognized \textit{BERT-base-uncased} model, which is accessible through the Hugging Face Model Hub\footnote{\url{https://huggingface.co/bert-base-uncased}}. This model serves as a crucial component in the process of deriving meaningful representations of the news articles, facilitating the assessment of their similarity in a comprehensive manner.

\textbf{SBERT} (Sentence-BERT) represents a modification of the pretrained BERT architecture~\cite{reimers2019sentence}.
This variant employs siamese and triplet network structures to produce semantically rich sentence embeddings. These embeddings are designed to be directly comparable using cosine similarity, enabling effective semantic similarity assessments between sentences. In our experiments, we employed the \textit{all-mpnet-base-v2} model\footnote{\url{https://huggingface.co/sentence-transformers/all-mpnet-base-v2}}.

\textbf{SimCSE}, is an innovative contrastive learning framework~\cite{gao2021simcse} that has consistently outperformed BERT-base on multiple Semantic Textual Similarity (STS) datasets, showcasing its robustness and effectiveness.

In our study, we employ \textit{bert-base-uncased} model as the encoder for training SimCSE. To ensure the fidelity and reliability of our experiments, we utilize the original implementation of SimCSE provided by the authors, which is available at the following repository\footnote{\url{https://github.com/princeton-nlp/SimCSE}}. This decision aligns with the best practices in the field and contributes to the credibility of our findings.

In the experimental phase involving BERT, SBERT, and SimCSE, we proceed by generating embeddings for each news article and subsequently calculated the cosine similarity between these embeddings. Our aim was to identify the optimal threshold value. Conversely, in the case of MinHash, we began by generating signatures for each news article, and our threshold selection process involved the application of the Jaccard similarity coefficient to determine the most suitable threshold value.

\subsection{Experiment Configurations}
We systematically explore a comprehensive range of thresholds, spanning from 0 to 1, with increments of 0.01. Our objective is to identify the threshold that yields the highest F1-score on the training dataset, as this threshold selection is critical for optimal performance.

Once the threshold is determined, we utilize it to calculate the F1-score on the testing dataset, ensuring that our evaluation is consistent with the chosen threshold.

These experiments are conducted using a dedicated setup, leveraging a single V100-32GB GPU for BERT, SBERT, SimCSE, and 72 CPU cores for Minhash computations. The computational intensity of these experiments is noteworthy, with each model requiring approximately 12 hours to complete computations across all levels and topics.

\subsection{Results}
We conducted comprehensive experiments using all the models previously described, evaluating them on datasets generated through four distinct approaches: Intra Topic, Inter Topic, Hard Examples, and Combined. The outcomes of the experiments are presented in Table ~\ref{tab2}.

\textbf{Inter-Topic.}
SBERT achieved the highest F1-score at 0.920, followed by SimCSE at 0.896, outperforming MinHash, which achieved an F1-score of 0.707. These results suggest that determining similarity within news pairs, particularly those created using the inter-topic approach, poses a greater challenge compared to identifying similar news when utilizing datasets from the other approaches.

\textbf{Intra-Topic.}
In intra-topic experiments, SimCSE achieved an F1-score of 0.890 in intra-topic evaluations, slightly lower than its inter-topic performance.
Meanwhile, MinHash, BERT, and SBERT demonstrated varying results between the two settings. MinHash achieved an F1-score of 0.760 in intra-topic experiments, BERT scored 0.842, and SBERT outperformed with an F1-score of 0.933 in intra-topic assessments, making it the top performer within the same thematic category.

\textbf{Hard Examples.}
In comparison to the intra-similarity approach, the hard examples approach yields lower results across all models. Among these models, SBERT demonstrates the strongest performance, achieving an impressive F1-score of 0.924. Following closely behind are SimCSE, BERT, and MinHash, with the latter displaying the lowest average performance among the group.

\textbf{Combined.}
In the context of the combined approach, it's worth noting that SBERT stands out with the highest F1-score, 0.922. This performance aligns closely with the outcomes seen in the other approaches we've explored. On the opposite end of the spectrum, MinHash delivers the lowest F1-score, registering at 0.757, and this pattern remains consistent across the various approaches.

For all four approaches to pair creation, MinHash demonstrated lower results than BERT, SBERT, and SimCSE, highlighting the advantage of using embeddings compared to the more classical MinHash approach. Additionally, SBERT achieved the highest F1-score in all cases, which may be attributed to our use of SBERT in the news filtering process. In second place was SimCSE, which showed better results than BERT for all types of pairs, confirming the advantage of using SimCSE over BERT on the Semantic Textual Similarity (STS) datasets~\cite{gao2021simcse}.

\section{Conclusion and Future Work}
In this paper, we propose a novel semantic textual similarity dataset for news data which accounts for emergent semantic categories within the text. While there are a wide variety of available textual similarity datasets, they fail to account for structural patterns that exist within text, which are generally easier for machine learning systems to classify.

We meticulously select and curate examples that pertain to distinct news topics, with the aim of creating more complex textual pairs for our study. This process led to the development of a total of 32 training and test datasets for news similarity detection. These datasets are organized based on four distinct approaches for generating news pairs: Inter-Topic Similarity, Intra-Topic Similarity, Hard Example Mining, and Combined Similarity.

The experimental results indicate that our dataset poses a significantly greater challenge for state-of-the-art models, underscoring the inherent difficulty of the task at hand. In the spirit of advancing research in this domain, we make this dataset readily accessible to the wider community. Our primary objective is to contribute valuable resources that can be instrumental in enhancing the performance and capabilities of future models.

In terms of future work, our primary objective is to expand the dataset's scope. We aim to transform it into a comprehensive resource, encompassing not only multiple topics but also multiple languages, thus fostering cross-lingual analysis and training/testing with news data in various domains.

Additionally, we intend to conduct a thorough comparative analysis, pitting our dataset against existing ones like SemEval-2022. This comparison will help us gauge the interchangeability and compatibility of these datasets.

\section{Ethics Statement}

The dataset used in this study comprises articles sourced exclusively from Wikinews, a publicly accessible news platform. The models employed in the dataset creation are likewise publicly available online. During the labeling process, annotators adhered to the detailed rules we specified. Consequently, it is important to note that the dataset presented here does not incorporate any confidential personal information.

The models used for benchmarking are likewise accessible online. As part of our unwavering dedication to transparency, we have made available the complete dataset we generated, along with the outcomes achieved by all the models used in the benchmarking process.

\nocite{*}
\section{Bibliographical References}\label{sec:reference}

\bibliographystyle{lrec-coling2024-natbib}

\end{document}